# Supervision-by-Registration: An Unsupervised Approach to Improve the Precision of Facial Landmark Detectors

*Xuanyi Dong[1], Shoou-I Yu[2], Xinshuo Weng[2], Shih-En Wei[2], Yi Yang[1], Yaser Sheikh[2]

xuanyi.dong@student.uts.edu.au; yi.yang@uts.edu.au
{shoou-i.yu,xinshuo.weng,shih-en.wei,yaser.sheikh}@fb.com

[1]CAI, University of Technology Sydney, [2]Facebook Reality Labs, Pittsburgh

## Abstract

*In this paper, we present supervision-by-registration, an unsupervised approach to improve the precision of facial landmark detectors on both images and video. Our key observation is that the detections of the same landmark in adjacent frames should be coherent with registration, i.e., optical flow. Interestingly, the coherency of optical flow is a source of supervision that does not require manual labeling, and can be leveraged during detector training. For example, we can enforce in the training loss function that a detected landmark at $frame_{t-1}$ followed by optical flow tracking from $frame_{t-1}$ to $frame_t$ should coincide with the location of the detection at $frame_t$. Essentially, supervision-by-registration augments the training loss function with a registration loss, thus training the detector to have output that is not only close to the annotations in labeled images, but also consistent with registration on large amounts of unlabeled videos. End-to-end training with the registration loss is made possible by a differentiable Lucas-Kanade operation, which computes optical flow registration in the forward pass, and back-propagates gradients that encourage temporal coherency in the detector. The output of our method is a more precise image-based facial landmark detector, which can be applied to single images or video. With supervision-by-registration, we demonstrate (1) improvements in facial landmark detection on both images (300W, ALFW) and video (300VW, Youtube-Celebrities), and (2) significant reduction of jittering in video detections.*

## 1. Introduction

Precise facial landmark detection lays the foundation for high quality performance of many computer vision and computer graphics tasks, such as face recognition [15], face animation [2] and face reenactment [33]. Many face recognition methods rely on locations of detected facial land-marks to spatially align faces, and imprecise landmarks could lead to bad alignment and degrade face recognition performance. In face animation and reenactment methods, 2D landmarks are used as anchors to deform 3D face meshes toward realistic facial performances, so temporal jittering of 2D facial landmark detections in video will be propagated to the 3D face mesh and could generate perceptually jarring results.

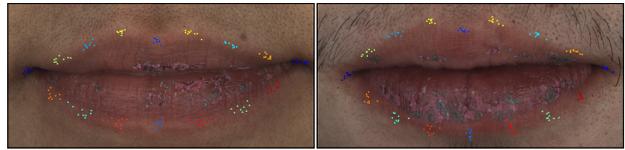

Figure 1. **Annotations are imprecise.** We show annotations of nine annotators on two images of the mouth. Each color indicates a different landmark. Note the inconsistencies of annotations even on the more discriminative landmarks such as the corner of the mouth. This could be harmful to both the training and evaluation of detectors, thus motivating the use of supervisory signals which does not rely on human annotations.

Precise facial landmark detection is still an unsolved problem. While significant work has been done on image-based facial landmark detection [19, 28, 38], these detectors tend to be accurate but not precise, i.e., the detector's bias is small but variance is large. The main causes could be: (1) insufficient training samples and (2) imprecise annotations, as human annotations inherently have limits on precision and consistency as shown in Figure 1. As a result, jittering is observed when we apply the detector independently to each video frame, and the detected landmark does not adhere well to an anatomically defined point (e.g., mouth corner) on the face across time. Other methods that focus on video facial landmark detection [13, 22, 23] utilize both detections and tracking to combat jittering and increase precision, but these methods require per-frame annotations in video, which are (1) tedious to annotate due to the sheer volume of video frames and (2) difficult to annotate consistently across frames, even for temporally adjacent frames.

---

*Work done during an internship at Facebook Oculus

Therefore, precise facial landmark detection might not be simply solved with large amounts of human annotations.

Instead of completely relying on human annotations, we present *Supervision-by-Registration* (SBR), which augments the training loss function with supervision automatically extracted from unlabeled videos. The key observation is that the coherency of (1) the detections of the same landmark in adjacent frames and (2) registration, i.e., optical flow [18], is a source of supervision. This supervision can complement the existing human annotations during the *training* of the detector. For example, a detected landmark at frame$_{t-1}$ followed by optical flow tracking between frame$_{t-1}$ and frame$_t$ should coincide with the location of the detection at frame$_t$. So, if the detections are incoherent with the optical flow, the amount of mismatch is a supervisory signal enforcing the detector to be temporally consistent across frames, thus enabling a SBR-trained detector to better locate the correct location of a landmark that is hard to annotate precisely. The key advantage of SBR is that no annotations are required, thus the training data is no longer constrained by the quantity and quality of human annotations.

The overview of our method is shown in Figure 2. Our end-to-end trainable model consists of two components: a generic detector built on convolutional networks [16], and a differentiable Lucas-Kanade (LK, [1, 4, 18]) operation. During the forward pass, the LK operation takes the landmark detections from the past frame and estimates their locations in the current frame. The tracked landmarks are then compared with the direct detections on the current frame. The registration loss is defined as the offset between them. In the backward pass, the gradient from the registration loss is back-propagated through the LK operation to encourage temporal coherency in the detector. To ensure that the supervision from registration is reasonable, supervision is only enforced for landmarks whose optical flow pass the forward-backward check [12]. The final output of our method is an enhanced image-based facial landmark detector which has leveraged large amounts of unlabeled video to achieve higher precision in both images and videos, and more stable predictions in videos.

Note that our approach is fundamentally different from post-processing such as temporal filtering, which often sacrifices precision for stability. Our method directly incorporates the supervision of temporal coherency during model *training*, thus producing detectors that are inherently more stable. Therefore, neither post-processing, optical flow tracking, nor recurrent units are required upon per-frame detection in test time. Also note that SBR is not regularization, which limits the freedom of model parameters to prevent overfitting. Instead, SBR brings more supervisory signals from registration to enhance the precision of the detector. In sum, SBR has the following benefits:

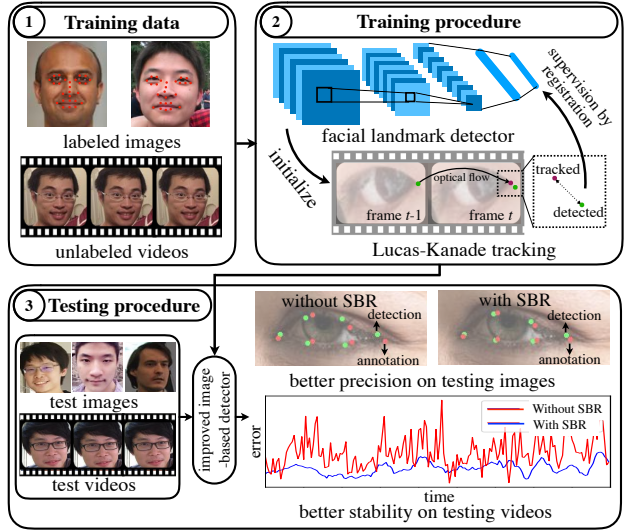

Figure 2. The **supervision-by-registration (SBR) framework** takes labeled images and unlabeled video as input to train an image-based facial landmark detector which is more precise on images/video and also more stable on video.

1. SBR can enhance the precision of a generic facial landmark detector on both images and video in an unsupervised fashion.
2. Since the supervisory signal of SBR does not come from annotations, SBR can utilize a very large amount of unlabeled video to enhance the detector.
3. SBR can be trained end-to-end with the widely used gradient back-propagation method.

## 2. Related Work

Facial landmark detection is mainly performed on two modalities: images and video. In images, the detector can only rely on the static image to detect landmarks, whereas in video the detector has additional temporal information to utilize. Though image-based facial landmark detectors [8, 19, 38, 39, 3, 20] can achieve very good performance on images, sequentially running these detectors on each frame of a video in a tracking-by-detection fashion usually leads to jittering and unstable detections.

There are various directions for improving facial landmark detection in videos apart from tracking-by-detection. Pure temporal tracking [1, 9] is a common method but often suffer from tracker drift. Once the tracker has failed in the current frame, it is difficult to make the correct prediction in the following frames. Therefore, hybrid methods [13, 17, 22] jointly utilize tracking-by-detection and temporal information in a single framework to predict more stable facial landmarks. Peng et al. [22] and Liu et al. [17] utilize recurrent neural networks to encode the temporal information across consecutive frames. Khan et al. [13] uti-lize global variable consensus optimization to jointly optimize detection and tracking in consecutive frames. Unfortunately, these methods require per-frame annotations, which are resource-intensive to acquire. Our approach SBR shares the high-level idea of these algorithms by leveraging temporal coherency, but SBR does not require any video-level annotation, and is therefore capable of enhancing detectors from large numbers of unlabeled videos.

Other approaches utilize temporal information in video to construct person-specific models [27, 23, 24]. Most of these methods usually leverage offline-trained static appearance models. The detector, which is used to generate initial landmark prediction, is not updated based on the tracking result in their algorithms, whereas SBR dynamically refines the detector based on LK tracking results. Self-training [42] can also be utilized for creating person-specific models, and was shown to be effective in pose estimation [5, 30]. However, unlike our method which can be trained end-to-end, [5, 30] did alternating bootstrapping to progressively improve the detectors. This leads to longer training times, and also inaccurate gradient updates as detailed in Sec. 5.

## 3. Methodology

SBR consists of two complementary parts, the general facial landmark detector and the LK tracking operation, as shown in Figure 3. The key idea of this framework is that we can directly perform back-propagation through the LK operation, thus enabling the detector before the LK operation to receive gradients which encourage temporal coherency across adjacent frames. LK was chosen because it is fully differentiable.

### 3.1. LK Operation

Motivated by [4], we design an LK operation through which we can perform back-propagation. Given the feature $\mathbf{F}_{t-1}$[1] from frame$_{t-1}$ and feature $\mathbf{F}_t$ from the frame$_t$, we estimate the parametric motion for a small patch near $\mathbf{x}_{t-1} = [x, y]^T$ from frame$_{t-1}$. The motion model is represented by the displacement warp function $W(\mathbf{x}; \mathbf{p})$. A displacement warp contains two parameters $\mathbf{p} = [p_1, p_2]^T$, and can be formulated as $W(\mathbf{x}; \mathbf{p}) = [x + p_1, y + p_2]^T$. We leverage the inverse compositional algorithm [1] for our LK operation. It finds the motion parameter $\mathbf{p}$ by minimizing

$$\sum_{\mathbf{x} \in \Omega} \alpha_{\mathbf{x}} \parallel \mathbf{F}_{t-1}(W(\mathbf{x}; \Delta\mathbf{p})) - \mathbf{F}_t(W(\mathbf{x}; \mathbf{p})) \parallel^2, \quad (1)$$

with respect to $\Delta\mathbf{p}$. Here, $\Omega$ is a set of locations in a patch centered at $\mathbf{x}_{t-1}$, and $\alpha_{\mathbf{x}} = \exp(-\frac{\|\mathbf{x}-\mathbf{x}_{t-1}\|_2^2}{2\sigma^2})$ is the weight value for $\mathbf{x}$ determined by the distance from $\mathbf{x}_{t-1}$ to down-weight pixels further away from the center of the patch. Af-

[1]The features can be RGB images or the output of convolution layers.

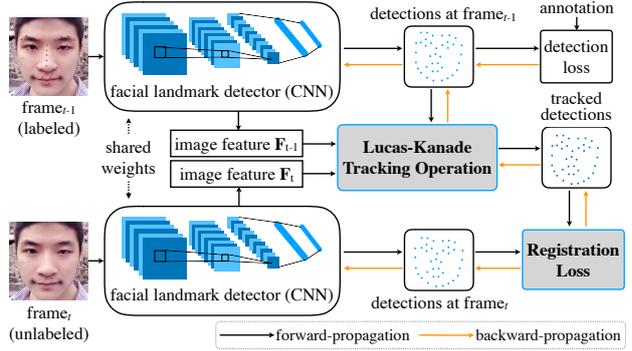

Figure 3. The **training procedure** of supervision-by-registration with two complementary losses. The detection loss utilizes appearance from a single image and label information to learn a better landmark detector. The registration loss uncovers temporal consistency by incorporating a Lucas-Kanade operation into the network. Gradients from the registration loss are back-propagated through the LK operation to the detector network, thus enforcing the predictions in neighboring frames to be consistent.

ter obtaining the motion parameter, the LK operation updates the warp parameter as follows:

$$W(\mathbf{x}; \mathbf{p}) \leftarrow W(W(\mathbf{x}; \Delta\mathbf{p})^{-1}; \mathbf{p}) = \left[ \begin{array}{c} x + p_1 - \Delta p_1 \\ y + p_2 - \Delta p_2 \end{array} \right]. \quad (2)$$

$\mathbf{p}$ is an initial motion parameter ($\mathbf{p} = [0, 0]^T$ in our case), which will be iteratively updated by Eq. (2) until convergence.

The first order Taylor expansion on Eq. (1) gives:

$$\sum_{\mathbf{x} \in \Omega} \alpha_{\mathbf{x}} \parallel \mathbf{F}_{t-1}(W(\mathbf{x}; \mathbf{0})) + \nabla \mathbf{F}_{t-1} \frac{\partial W}{\partial \mathbf{p}} \Delta\mathbf{p} - \mathbf{F}_t(W(\mathbf{x}; \mathbf{p})) \parallel^2 \quad (3)$$

We then have the solution to Eq. (3) according to [1]:

$$\Delta\mathbf{p} = \mathbf{H}^{-1} \sum_{\mathbf{x} \in \Omega} J(\mathbf{x})^T \alpha_{\mathbf{x}} (\mathbf{F}_t(W(\mathbf{x}; \mathbf{p})) - \mathbf{F}_{t-1}(W(\mathbf{x}; \mathbf{0}))), \quad (4)$$

where $\mathbf{H} = \mathbf{J}^T \mathbf{A} \mathbf{J} \in \mathbb{R}^{2 \times 2}$ is the Hessian matrix. $\mathbf{J} \in \mathbb{R}^{C|\Omega| \times 2}$ is the vertical concatenation of $J(\mathbf{x}) \in \mathbb{R}^{C \times 2}$, $\mathbf{x} \in \Omega$, which is the Jacobian matrix of $\mathbf{F}_{t-1}(W(\mathbf{x}; \mathbf{0}))$. $C$ is the number of channels of $\mathbf{F}$. $\mathbf{A}$ is a diagonal matrix, where elements in the main diagonal are the $\alpha_{\mathbf{x}}$'s corresponding to the $\mathbf{x}$'s used to create $\mathbf{J}$. $\mathbf{H}$ and $\mathbf{J}$ are constant over iterations and can thus be pre-computed.

We illustrate the detailed steps of the LK operation in Figure 4, and describe it in Algorithm 1. We define the LK operation as $\tilde{\mathbf{L}}_t = G(\mathbf{F}_{t-1}, \mathbf{F}_t, \mathbf{L}_{t-1})$. This function takes a matrix $\mathbf{L}_{t-1} = [\mathbf{x}_{t-1}^1, \mathbf{x}_{t-1}^2, ..., \mathbf{x}_{t-1}^K] \in \mathbb{R}^{2 \times K}$, which represents the coordinates of $K$ landmarks from frame$_{t-1}$, as input to generate the landmarks $\tilde{\mathbf{L}}_t$ for the next (future) frame. Since, all steps in the LK operation are differentiable, the gradient can back-propagate to the facial landmark locations and the feature maps through LK.

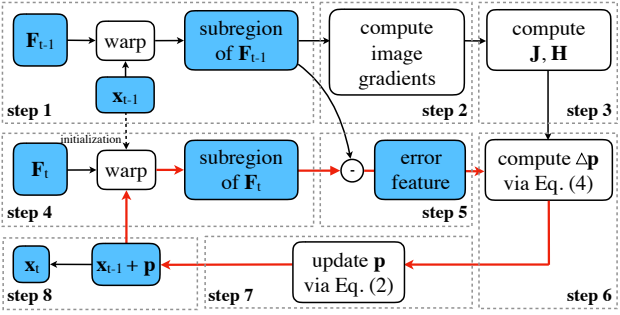

Figure 4. **Overview of the LK operation.** This operation takes the features of two adjacent frames, i.e., $\mathbf{F}_{t-1}$ and $\mathbf{F}_t$, and a location $\mathbf{x}_{t-1}$ at frame$_{t-1}$ as inputs. The inverse compositional LK algorithm iteratively updates the motion parameter $\mathbf{p}$ and outputs the corresponding coordinates $\mathbf{x}_t$ at frame$_t$. The iterative portion of the algorithm is indicated by the red arrows. Every step of this process is differentiable, thus gradients can back-propagate from $\mathbf{x}_t$ to $\mathbf{F}_{t-1}$, $\mathbf{F}_t$ and $\mathbf{x}_{t-1}$.

We apply a very small value to the diagonal elements of $\mathbf{H}$. This ensures that $\mathbf{H}$ is invertible. Also, in order to crop a patch at a sub-pixel location $\mathbf{x}$, we use the spatial transformer network [11] to calculate the bilinear interpolated values of the feature maps.

### 3.2. Supervision-by-Registration

We describe the details of the two complementary losses: the detection loss based on human annotations and the registration loss used to enforce temporal coherency.

**Detection loss.** Many facial landmark detectors take an image $\mathbf{I}$ as input and regresses to the coordinates of the facial landmarks, i.e., $D(\mathbf{I}) = \mathbf{L}$. They usually apply an $L_2$ loss on these coordinates $\mathbf{L}$ with the ground-truth labels $\mathbf{L}^*$, i.e., $\ell_{\text{det}} = ||\mathbf{L} - \mathbf{L}^*||_2^2$.

Other methods [36, 20] predict a heat-map rather than the coordinates for each landmark, and the $L_2$ loss is usually applied on the heatmap during the training procedure. During testing, instead of directly regressing the location of the landmarks, the argmax operation is used on the heatmap to obtain the location of the landmark. Unfortunately, the argmax operation is not differentiable, so these methods cannot be directly used with our LK operation. To enable the information to be back-propagated through the predicted coordinates in heatmap-based methods, we replace the argmax operation with a heatmap peak finding operation which is based on a weighted sum of heatmap confidence scores. Let $\mathbf{M}$ be the predicted heatmap. For each landmark, we first compute a coarse location using $\arg\max \mathbf{M} = [x', y']^T$. Then we crop a small square region with edge length $2 \times r$ centered at $[x', y']^T$, denoted as

**Algorithm 1** Algorithm Description of the LK operation
**Input:** $\mathbf{F}_{t-1}, \mathbf{F}_t, \mathbf{x}_{t-1}, \mathbf{p} = [0, 0]^T$
 1. Extract template feature from $\mathbf{F}_{t-1}$ centered at $\mathbf{x}_{t-1}$
 2. Calculate the gradient of the template feature
 3. Pre-compute the Jacobian and Hessian matrices, $\mathbf{J}$ and $\mathbf{H}$
 **for** $iter = 1; iter \leq max$; iter++ **do**
  4. Extract target feature from $\mathbf{F}_t$ centered at $\mathbf{x}_{t-1} + \mathbf{p}$
  5. Compute the error of the template and target features
  6. Compute $\Delta \mathbf{p}$ using Eq. (4)
  7. Update the motion model $\mathbf{p}$ using Eq. (2)
 **end for**
**Output:** $\mathbf{x}_t = \mathbf{x}_{t-1} + \mathbf{p}$

$\mathcal{M}' = \{(i, j) \mid i \in [x' - r, x' + r], j \in [y' - r, y' + r]\}$. Lastly, we use the soft-argmax operation on this square region to obtain the final coordinates:

$$\mathbf{x} = \frac{\sum_{(i,j) \in \mathcal{M}'} \mathbf{M}_{i,j} \times [i, j]^T}{\sum_{(i,j) \in \mathcal{M}'} \mathbf{M}_{i,j}}. \quad (5)$$

Since Eq. (5) is differentiable, we can utilize this peak finding operation to incorporate heatmap-based methods into our framework. Note that when we train with different kinds of networks, we can still use the original loss functions and settings described in their papers [36, 20, 19, 41], but for simplicity, we still denote in this paper the detection loss as the $L_2$ distance between the predicted and ground-truth coordinates.

**Registration loss.** Registration loss can be computed in an unsupervised manner to enhance the detector. It is realized with a forward-backward communication scheme between the detection output and the LK operation, as shown in Figure 5. The forward communication computes the registration loss while the backward communication evaluates the reliability of the LK operation.

In the forward communication, the detector passes the detected landmarks of past frames (e.g., $\mathbf{L}_{t-1}$ for frame$_{t-1}$) to the LK operation. The LK operation then generates new landmarks of future frames (e.g., $\tilde{\mathbf{L}}_t = G(\mathbf{F}_{t-1}, \mathbf{F}_t, \mathbf{L}_{t-1})$ for frame$_t$) by tracking. LK-generated landmarks for future frames should be spatially near the detections in the future frames. Therefore, the registration loss directly computes the distance between the LK operation's predictions (green dots in Figure 5) and the detector's predictions (blue dots in Figure 5), thus encouraging the detector to be more temporally consistent. The loss is as follows:

$$\ell_{\text{regi}}^t = \sum_{i=1}^{K} \beta_{t,i} ||\mathbf{L}_{t,i} - \tilde{\mathbf{L}}_{t,i}||_2 \quad (6)$$
$$= \sum_{i=1}^{K} \beta_{t,i} ||\mathbf{L}_{t,i} - G(\mathbf{F}_{t-1}, \mathbf{F}_t, \mathbf{L}_{t-1,i})||_2.$$

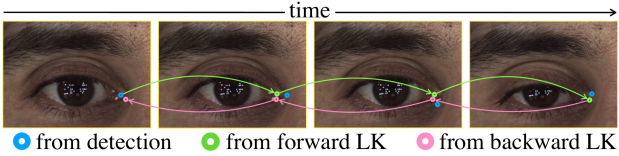

○ from detection　○ from forward LK　○ from backward LK

Figure 5. **Forward-backward communication scheme** between the detector and the LK operation during the training procedure. The green and pink lines indicate the forward and backward LK tracking routes. The blue/green/pink dots indicate the landmark predictions from the detector/forward-LK/backward-LK. The forward direction of this communication adjusts the detection results of future frames based on the past frame. The backward direction assesses the reliability of the LK operation output.

$\mathbf{L}_{t,i}$ and $\tilde{\mathbf{L}}_{t,i}$ denote the $i$-th row of $\mathbf{L}_t$ and $\tilde{\mathbf{L}}_t$, which correspond to the $i$-th landmark location. $\beta_{t,i} \in \{0, 1\}$ indicates the reliability of the $i$-th tracked landmark at time $t$, which is determined by the backward communication scheme.

The LK tracking may not always succeed, and supervision should not be applied when LK tracking fails. Therefore, the backward communication stream utilizes the forward-backward check [12] to evaluate the reliability of LK tracking. Specifically, the LK operation takes $\tilde{\mathbf{L}}_t$ as input and generates the landmarks of frame$_{t-1}$ by tracking in reverse order, formulated as: $\hat{\mathbf{L}}_{t-1} = G(\mathbf{F}_t, \mathbf{F}_{t-1}, \tilde{\mathbf{L}}_t)$. Our premise is that if the LK operation output is reliable, a landmark should return to the same location after forward-backward tracking. Therefore, if the backward tracks are reliable, then $\beta_{t,i} = 1$ else $\beta_{t,i} = 0$, i.e., this point is not included in the registration loss. Since only reliable tracks will be used, the forward-backward communication scheme ensures that the registration loss yields improvement in performance when unlabeled data are exploited. Note that the registration loss is not limited to adjacent frames and can be applied to a sequence of frames as shown in Figure 5.

**Complete loss function.** Let $N$ be the number of training samples with ground truth. For notation brevity, we assume there is only one unlabeled video with $T$ frames. Then, the complete loss function of SBR is as follows:

$$\ell_{\text{final}} = \sum_{n=1}^{N} \ell_{\text{det}}^n + \gamma \sum_{t=1}^{T-1} \ell_{\text{regi}}^t, \quad (7)$$

which is a weighted combination of the detection and registration loss controlled by the weight parameter $\gamma$.

**Computation Complexity.** The computational cost of the LK module consists of two parts, the pre-computed operations, and iterative updating. The cost of the first part is $O(C|\Omega|)$, where $C$ is the channel size of the input ($C = 3$ for RGB images), and $|\Omega|$ is the patch size used in LK which is usually less than $10 \times 10$. The second part is $O(TC|\Omega|)$, where $T$ is the number of iterations and usually less than 20. Therefore, for all $K$ landmarks, the LK cost is $O(KC|\Omega|) + O(KCT|\Omega|)$, which is negligible when compared to the complexity of evaluating a CNN.

### 3.3. Personalized Adaptation Modeling

SBR can also be used to generate personalized facial landmark detectors, which is useful in (1) unsupervised adaptation to a testing video which may be in a slightly different domain than the training set and (2) generating the best possible detector for a specific person, e.g., a star actor in a movie. SBR can achieve this by treating testing videos as unlabeled videos and including them in training. During the training process, the detector can remember certain personalized details in an unsupervised fashion to achieve more precise and stable facial landmark detection.

## 4. Evaluation and Results

### 4.1. Datasets

**300-W** [26, 28, 29] provides annotations for 3837 face images with 68 landmarks. We follow [40, 19, 41] to split the dataset into four sets, training, common testing, challenging testing, and full testing, respectively.

**AFLW** [15] consists in total of 25993 faces in 21997 real-world images, where each face is annotated with up to 21 landmarks. Following [19], we ignore two landmarks of ears and only use the remaining 19 landmarks.

**YouTube-Face** [37] contains 3425 short videos of 1595 different people. This dataset does not have facial landmark labels, but the large variety of people makes it very suitable to provide to SBR as unlabeled video. We filter videos with low resolution[2], and use the remaining videos to train SBR in an unsupervised way.

**300-VW** [6, 31, 35]. This video dataset contains 50 training videos with 95192 frames. The test set consists of three categories with different levels. These three subsets (A, B and C) have 62135, 32805 and 26338 frames, respectively. C is the most challenging one. Following [13], we report the results for the 49 inner points on subset A and C.

**YouTube Celebrities** [14]. This dataset contains videos of 35 celebrities under varying poses, illumination and occlusion. Following the same setting as in [23], we perform PAM on the same six video clips as [23].

### 4.2. Experiment Settings

**Baselines.** We exploit two facial landmark detectors as baselines on which we further perform SBR and PAM.

The first detector is CPM [36], which utilizes the ImageNet pre-trained models [32, 7, 10] as the feature extraction part. In our experiment, we use the first four convolutional layers of VGG-16 [32] for feature extraction and use only three CPM stages for heatmap prediction. The faces are cropped and resized into 256×256 for pre-processing. We train the CPM with a batch size of 8 for 40 epochs in total. The learning rate starts at 0.00005 and is reduced by 0.5 at 20th and 30th epochs.

---
[2]Videos with mean face size $< 100^2$ are considered as low-resolution.

| Method | 300-W | | | AFLW |
|---|---|---|---|---|
| | Common | Challenging | Full Set | |
| SDM [38] | 5.57 | 15.40 | 7.52 | 5.43 |
| LBF [25] | 4.95 | 11.98 | 6.32 | 4.25 |
| MDM [34] | 4.83 | 10.14 | 5.88 | - |
| TCDCN [39] | 4.80 | 8.60 | 5.54 | - |
| CFSS [40] | 4.73 | 9.98 | 5.76 | 3.92 |
| Two-Stage [19] | 4.36 | **7.56** | 4.99 | 2.17 |
| Reg | 8.14 | 16.90 | 9.85 | 5.01 |
| Reg + SBR | 7.93 | 15.98 | 9.46 | 4.77 |
| CPM | 3.39 | 8.14 | 4.36 | 2.33 |
| CPM + SBR | **3.28** | 7.58 | **4.10** | **2.14** |

Table 1. Comparison of NME on 300-W and AFLW datasets.

The second detector is a simple regression network, denoted as Reg. We use VGG-16 as our base model and change the output neurons of the last fully-connected layer to $K \times 2$, where $K$ is the number of landmarks. Since VGG-16 requires the input size to be 224×224, we thus resize the cropped face to 224×224 for this regression network. Following [17], we normalize the $L_2$ loss by the size of faces as the detection loss.

**Training with LK.** We perform LK tracking over three consecutive frames. For $\Omega$ in Eq. (1), we crop a $10 \times 10$ patch centered at the landmark. To cope with faces with different resolutions, we resize the images accordingly such that a $10 \times 10$ crop is a reasonable patch size. Too large or small patch size can lead to poor LK tracking and hurt performance. The maximum iterations of LK is 20 and the convergence threshold for $\Delta \mathbf{p} = 10^{-6}$. For the input feature of the LK operation, we use the RGB image by default and also perform ablation studies when using the conv-1 feature layer (see Section 5). The weight of the registration loss is $\gamma = 0.5$. When training a model from scratch, we first make sure the detection loss has converged before activating the registration loss. When training with SBR, the ratio of labeled images and unlabeled video for each batch should be balanced. In the case when there are more unlabeled video than labeled images, we duplicate the labeled images such that the ratio is still balanced. Also, when applying SBR, one should confirm that the distribution of faces in unlabeled video is similar to the distribution of labeled images. Otherwise, the initial detector may perform poorly on the unlabeled videos, which leads to very few reliable LK tracks and a less effective PAM. All of our experiments are implemented in PyTorch [21].

**Evaluation Metrics.** Normalized Mean Error (NME) is used to evaluate the performance on images. Following [19, 8, 25], the interocular distance and face size is employed to normalize mean error on 300-W and AFLW respectively. We also use Cumulative Error Distribution (CED) [26] and Area Under the Curve (AUC) [13] for evaluation.

| Method | DGCM [13] | CPM | CPM+SBR | CPM+SBR+PAM |
|---|---|---|---|---|
| AUC@0.08 | 59.38 | 57.25 | 58.22 | **59.39** |

Table 2. AUC @ 0.08 error on 300-VW category C. Note that SBR and PAM do not utilize any additional annotations, but can still improve the baseline CPM and achieve the state-of-the-art results.

| Method | SDM [38] | ESR [3] | RLB [25] | PIEFA [23] |
|---|---|---|---|---|
| NME | 5.85 | 5.61 | 5.37 | 4.92 |
| Ours | Reg | Reg+PAM | CPM | CPM+PAM |
| NME | 10.21 | 9.31 | 5.26 | **4.74** |

Table 3. Comparisons of NME on YouTube Celebrities dataset.

### 4.3. Evaluation on Image Datasets

In order to show that the proposed SBR can enhance generic landmark detectors, we show the results of SBR performed on both the Reg (regression-based) and CPM (heatmap-based) on AFLW and 300-W. We also compare against nine facial landmark detection algorithms.

**Results on 300-W.** As shown in Table 1, the baseline CPM obtains competitive performance (4.36 NME) on the full testing set of 300-W. We then run SBR with unlabeled videos from YouTube-Face for both the CPM and Reg, which further improves the CPM by a relative 7% and the Reg by a relative 6% without using any additional annotation. The compared results are provided by the official 300-W organizer [26, 28].

**Results on AFLW.** The distribution of face size on AFLW dataset is different from that of 300-W. Thus we resize the face to 176×176. Table 1 shows that SBR improves the CPM by a relative 9% and Reg by a relative 5%.

Overall, the SBR improves both CPMs and regression networks on 300-W and AFLW. We also achieve the state-of-the-art performance with CPMs. This demonstrates the flexibility and effectiveness of SBR. YouTube-Face is used as unlabeled videos in our experiments, but in hindsight, it may not be the best choice to enhance the detector on 300-W, because the size of faces in YouTube-Face is smaller than 300-W, and compression artifacts further affect LK tracking. By using a video dataset with higher resolution, our approach can potentially obtain higher performance.

### 4.4. Evaluation on Video Datasets

To show that SBR/PAM can enhance a detector to produce coherent predictions across frames, we evaluate on 300VW. We follow [13] and use the full training set. Since we lack images from the XM2VTS and FRGC datasets, we use the same number labeled images from an internal dataset instead of these two datasets. The images in 300-VW have a lower resolution than 300-W, thus we resize the images to 172×172 during training according to the face size statistics.

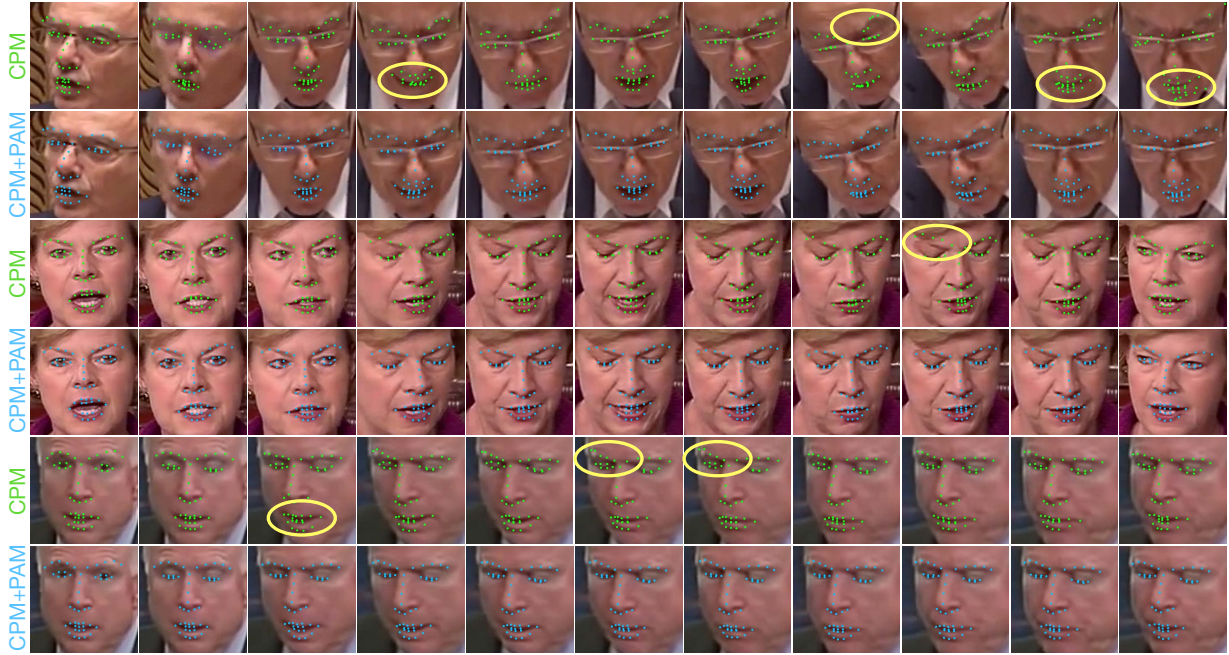

Figure 6. **Qualitative results** of CPM (green) and CPM+SBR/PAM (blue) on 300VW. We sample predictions every 10 frames from videos. Yellow circles indicate the clear failures from the CPM. CPM+SBR/PAM can produce more stable predictions across adjacent frames.

**Results on 300VW.** Table 2 shows that SBR improves the CPM by 1%, and PAM further improves it by 1.2%. t-Test shows a p-value of 0.0316 and 0.0001 when using SBR to enhance CPM and using PAM to enhance CPM+SBR. These two statistical significance tests demonstrate the improvement of SBR and PAM. We show that CPM+SBR+PAM achieves the state-of-the-art performance against all other methods. Importantly, SBR+PAM does not utilize any more annotations than what the baselines use.

**Results on YouTube Celebrities.** We also compare different personalized methods in Table 3. The baselines Reg and CPM are pre-trained on 300-W. The proposed PAM reduces the error of CPM from 5.26 to 4.74, achieving state-of-the-art performance.

**Qualitative comparison.** Figure 6 shows the qualitative results. CPM predictions are often incoherent across frames. For example, in the third row, the predictions on the eyebrow drifts, but SBR/PAM can produce more stable predictions as the coherency is satisfied during training.

## 5. Discussion

**Image Resolution.** The resolution of the face can affect the chance of the success of the LK operation as well as detector performance. Figure 7 shows that videos with higher face resolution usually result in a higher possibility to pass the forward-backward check. However, the perfor-mance improvement is not related to the face size. There could be other factors which have more influence, such as occlusion and head pose.

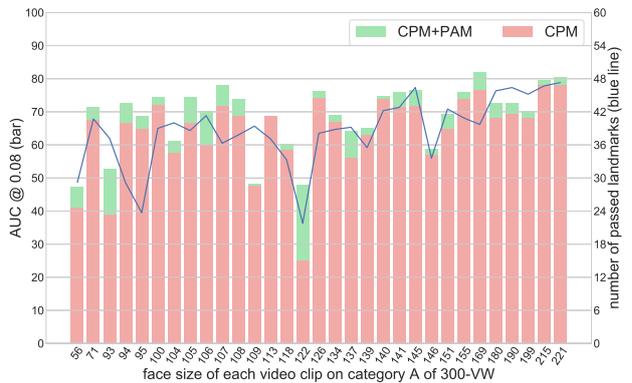

Figure 7. Analysis on category A of 300-VW. The x-axis indicates the face size of 31 videos in ascending order. The left y-axis indicates the AUC@0.08, and the right shows the number of landmarks that are considered as reliable by the forward-backward communication scheme.

**Temporal Length for Tracking.** The duration of LK tracking could be more than three consecutive frames. However, a longer period will result in a stricter forward-backward check, which reduces the number of landmarks to be included in the registration loss. We tested CPM + PAM with 5 frames LK tracking on YouTube Celebrities and achieved 5.01 NME, which is worse than the result of three frames, 4.74 NME.

**Image Features for Tracking.** We also tested using the conv-1 feature instead of the RGB image to do LK tracking, which resulted in an increase of error on YouTube Celebrities from 4.74 to 5.13 NME. This could be caused by the convolutional feature losing certain information that is useful for LK tracking, and more attention is required to learn features suitable for LK tracking.

**Effect of imprecise annotation.** SBR and PAM only show a small improvement based on the NME and AUC evaluation metric, but we observe significant reduction of jittering in videos (see demo video). There could be two reasons: (1) NME and AUC treat the annotations of each frame independently and does not take into account the smoothness of the detections, and (2) imprecise annotations in the testing set may adversely affect the evaluation results. We further analyze reason (2) by generating a synthetic face dataset named "SyntheticFace" from a 3D face avatar. The key advantage of a synthetic data set is that there is zero annotation error because we know exactly where each 3D vertex is projected into a 2D image. This enables us to analyze the effect of annotation errors by synthetically adding noise to the perfect "annotations". We generated 2537 training and 2526 testing face images under different expressions and identified 20 landmarks to detect. The image size is 5120×3840. We add varying levels of Gaussian noise to the training and testing set, which are then used to train and evaluate our detector. If we train on different levels of noise and evaluate the models on clean annotations, the testing performance is surprisingly close across models, as shown in Figure 8a. This means that our detector is able to "average out" the errors in annotation. However, the same models evaluated against testing annotations with varying error (Figure 8b) look significantly worse than Figure 8a. This means that a well-performing model may have poor results *simply due to the annotation error in the testing data*. In sum, annotation errors could greatly affect quantitative results, and a lower score does not necessarily mean no improvement.

**Connection with Self-Training.** Our method is interestingly a generalization of self-training, which was utilized by [5, 30] to take advantage of unlabeled videos in the pose estimation task. The procedure of self-training is (1) train a classifier with the current training set, (2) predict on unlabeled data, (3) treat high confidence predictions as pseudo-labels and add them to the training set, and (4) repeat step 1. The main drawback of this method is that high-confidence pseudo-labels are assumed "correct" and no feedback is provided to fix errors in the pseudo-labels.

In our problem setting, the pseudo-labels are $\tilde{\mathbf{L}}_t$, which are detections tracked from frame$_{t-1}$ with LK. If we simply perform self-training, $\tilde{\mathbf{L}}_t$ are directly used as labels to learn $\mathbf{L}_t$. No feedback to $\tilde{\mathbf{L}}_t$ is provided even if it is erro-

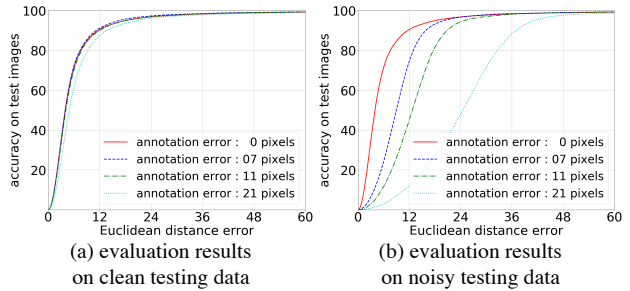

(a) evaluation results on clean testing data  (b) evaluation results on noisy testing data

Figure 8. **Effect of annotation error.** We add Gaussian noise to the annotations of SyntheticFace, and train the model on these noisy data. Different levels of Gaussian noise is indicated by different colors. **Left:** The models are evaluated on clean testing data of SyntheticFace. **Right:** The models are evaluated on noisy testing data of SyntheticFace, which has the same noise distribution as training.

neous. However, our registration loss provides feedback for both $\mathbf{L}_t$ and $\tilde{\mathbf{L}}_t$, thus if the pseudo-labels are inaccurate, $\tilde{\mathbf{L}}_t$ will also be adjusted. This is the key difference between our method and self-training. More formally, the gradient of our registration loss Eq. (6) with respect to the detector parameter $\theta$ is as follows:

$$\nabla_\theta \ell_{\text{regi}}^t = \sum_{i=1}^{K} \eta_{t,i}(\nabla_\theta \mathbf{L}_{t,i} - \nabla_\theta LK(\mathbf{F}_{t-1}, \mathbf{F}_t, \mathbf{L}_{t-1,i})),$$

where $\eta_{t,i} = \frac{\beta_{t,i}}{2||\mathbf{L}_{t,i}-LK(\mathbf{F}_{t-1},\mathbf{F}_t,\mathbf{L}_{t-1,i})||_2}$. For self-training, the gradients from $\tilde{\mathbf{L}}_t$: $\nabla_\theta LK(\mathbf{F}_{t-1}, \mathbf{F}_t, \mathbf{L}_{t-1,i})$ are missing. This compromises the correctness of the gradient for $\theta$, which is used to generate both $\mathbf{L}_t$ and $\tilde{\mathbf{L}}_t$. Empirically we observed that the detector tends to drift in a certain incorrect direction when the gradients of $\tilde{\mathbf{L}}_t$ are ignored, which led to an increase of error from 4.74 to 5.45 NME on YouTube Celebrities.

## 6. Conclusion

We present supervision-by-registration (SBR), which is advantageous because: (1) it does not rely on human annotations which tend to be imprecise, (2) the detector is no longer limited to the quantity and quality of human annotations, and (3) back-propagating through the LK layer enables more accurate gradient updates than self-training. Also, experiments on synthetic data show that annotation errors in the evaluation set may make a well-performing model seem like it is performing poorly, so one should be careful of annotation imprecision when interpreting quantitative results.

**Acknowledgment.** Yi Yang is partially supported by the Google Faculty Research Award, as well as the Data to Decisions CRC (D2D CRC) and the Cooperative Research Centres Programme.

10